\documentclass[letterpaper, 10 pt, conference]{ieeeconf}  
\IEEEoverridecommandlockouts          
\usepackage{amsmath,amsfonts}
\usepackage{algorithmic}
\usepackage{array}
\usepackage[caption=false,font=normalsize,labelfont=sf,textfont=sf]{subfig}
\usepackage{textcomp}
\usepackage{stfloats}
\usepackage{url}
\usepackage{verbatim}
\usepackage{graphicx}
\usepackage{multirow}
\usepackage{booktabs}
\usepackage{caption}
\usepackage{lipsum}
\usepackage{balance}
\usepackage{hyperref}
\usepackage{CJKutf8}
\usepackage{cite}
\usepackage{url}
\hypersetup{
	colorlinks=true,            
	linkcolor=red,              
	citecolor=green,            
	filecolor=blue,   
	urlcolor=blue            
}

\usepackage{url}
\usepackage{soul}

\usepackage{graphicx} 
\usepackage{amssymb}
\usepackage{makecell}

\usepackage{wrapfig}
\usepackage{booktabs}
\usepackage{tikz}
\usepackage{circledsteps}
\let\labelindent\relax
\usepackage{enumitem}

\begin{document}

\title{CoCMT: Communication-Efficient Cross-Modal Transformer \\ for Collaborative Perception}

\author{Rujia Wang$^{1}$, Xiangbo Gao$^{1}$, Hao Xiang$^{2}$, Runsheng Xu$^{2}$, Zhengzhong Tu$^{1*}$%
\thanks{ 
R. Wang, X. Gao, and Z. Tu are with Computer Science and Engineering, Texas A\&M University, College Station, TX 77840, USA.
H. Xiang and R. Xu are with University of California, Los Angeles, Los Angeles, CA 90095, USA.
* Corresponding author: Zhengzhong Tu ({\tt\small tzz@tamu.edu})}%
}


\maketitle

\begin{abstract}
Multi-agent collaborative perception enhances each agent’s perceptual capabilities by sharing sensing information to cooperatively perform robot perception tasks. This approach has proven effective in addressing challenges such as sensor deficiencies, occlusions, and long-range perception.
However, existing representative collaborative perception systems transmit intermediate feature maps, such as bird’s-eye view (BEV) representations, which contain a significant amount of non-critical information, leading to high communication bandwidth requirements.
To enhance communication efficiency while preserving perception capability, we introduce CoCMT, an object-query-based collaboration framework that optimizes communication bandwidth by selectively extracting and transmitting essential features.
Within CoCMT, we introduce the Efficient Query Transformer (EQFormer) to effectively fuse multi-agent object queries and implement a synergistic deep supervision to enhance the positive reinforcement between stages, leading to improved overall performance.
Experiments on OPV2V and V2V4Real datasets show CoCMT outperforms state-of-the-art methods while drastically reducing communication needs. 
On V2V4Real, our model (Top-50 object queries) requires only 0.416 Mb bandwidth—83 times less than SOTA methods—while improving AP@70 by 1.1\%. This efficiency breakthrough enables practical collaborative perception deployment in bandwidth-constrained environments without sacrificing detection accuracy. 
The code and models are open-sourced through the following link: \href{https://github.com/taco-group/COCMT}{https://github.com/taco-group/COCMT}.

\end{abstract}

\section{Introduction}
\label{sec:intro}

Accurate and efficient perception is essential for autonomous driving to ensure reliable navigation and safe decision-making~\cite{xing2024autotrust, xing2025openemma, wang2025generative, yang2025review}. However, single-vehicle autonomous systems face significant challenges in real-world scenarios, including occlusions and limited sensing range. Cooperative perception systems address these issues by enabling agents to enhance their perceptual capabilities through sharing sensing information with other agents.

\begin{figure}[ht]
    \centering
    
    \includegraphics[width=\columnwidth]{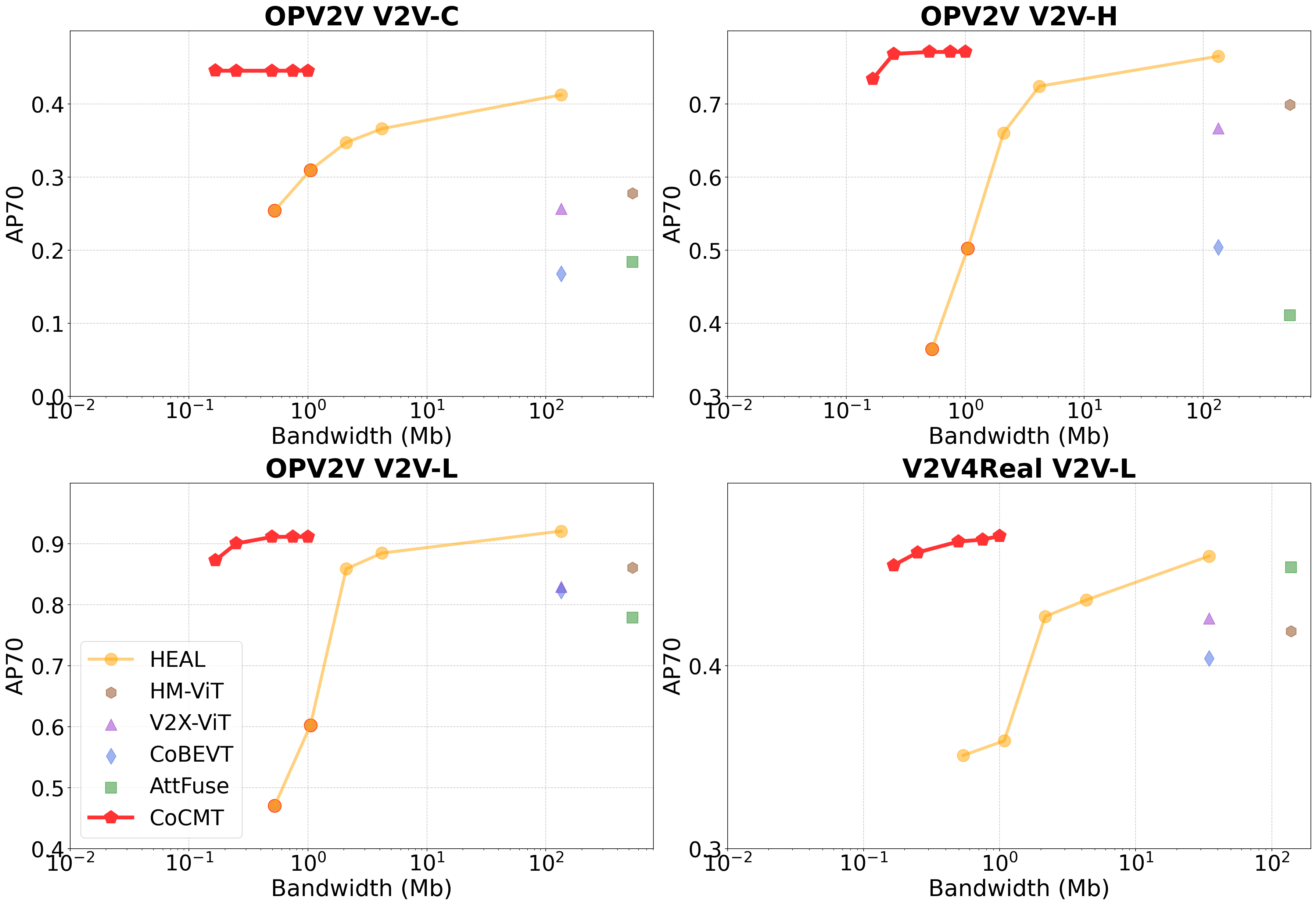}
    \caption{\textbf{AP70 vs Bandwidth under various communication bandwidth conditions}. We evaluated three modality settings on the OPV2V~\cite{OPV2VATTFuse} dataset: V2V-C, V2V-L, and V2V-H. Our CoCMT model, with a significant bandwidth advantage, demonstrates performance comparable to or even better than state-of-the-art methods, HEAL~\cite{HEAL}. Moreover, as the bandwidth exponentially decreases, CoCMT exhibits only minor performance degradation, fully showcasing its excellent adaptability to bandwidth fluctuations.} 
    \label{fig:opv2v_3branch_apvsbandwidth}
    \vspace{-5mm}
\end{figure}

Most existing cooperative perception fusion methods~\cite{OPV2VATTFuse, V2XViT, CoBEVFlow, gao2025stamp} use feature maps---such as Bird's-Eye-View (BEV) features---as the medium for information transmission among agents. Feature maps represent the entire scene surrounding the vehicle, where dynamic, relatively sparse foreground objects are mixed with a large amount of static background information. 
Transmitting large amounts of background data offers minimal benefit to perception performance while occupying significant bandwidth.
Some works improve communication efficiency but introduce other challenges, such as computational latency~\cite{Where2comm, hu2024pragmatic, CodeFilling} or being limited to only two agents~\cite{QUEST}.

In this paper, we introduce a novel object-centric framework tailored for communication-efficient collaborative perception.
The sparsity nature of query-based object representations~\cite{carion2020end, DN-DETR} has offered several advantages over prior feature map-based strategies:
\Circled{1} \underline{Small data size}: The data size of object queries is significantly smaller than that of the entire BEV feature maps, which can largely reduce the communication bandwidth.
 \Circled{2} \underline{Object-centric focus}: Unlike feature maps, object queries are explicitly \emph{object-centric}, encapsulating only the relevant contextual features and naturally excluding irrelevant background data. This eliminates the need for intermediate fusion algorithms to design complex foreground information extraction mechanisms~\cite{Where2comm, hu2024pragmatic, CodeFilling} and reduces the computational latency.
\Circled{3} \underline{Modality independence}: Object queries are less dependent on specific data modalities, making them more versatile and effective for agents with different sensor modalities.
These advantages make object queries a more efficient and scalable choice for cooperative perception systems, especially in bandwidth-constrained and multi-modal environments.

However, integrating object queries from multiple agents presents two key challenges. First, object queries are unordered, with adjacent queries potentially representing spatially distant objects. This unstructured nature causes feature confusion when merging information from three or more agents~\cite{QUEST}, complicating relevant object interactions. Second, object query-based models generate numerous initial queries, many unrelated to actual objects. Efficiently filtering these noisy queries to focus only on high-quality candidates is critical for effective fusion. To address these challenges, we propose the Efficient Query Transformer (EQFormer), which dynamically identifies important queries and eliminates noise, enabling more efficient and effective cooperative perception. We summarize the contribution of this work as follow:
\begin{itemize}[leftmargin=*]
    \item We propose CoCMT, a novel object query-based collaborative perception framework that uses object queries as intermediaries for information transmission, significantly reducing bandwidth consumption while enhancing the efficiency of collaborative perception.

    \item We design the Efficient Query Transformer (EQFormer), which selectively processes queries and eliminates noise based on the query validity, spatially proximate, and prediction confidence, ensuring focused and efficient attention learning for fusion.
    
    \item Our extensive experiments on the OPV2V and V2V4real datasets validate the bandwidth efficiency of our proposed framework. The results demonstrate that the framework significantly reduces bandwidth consumption while achieving superior performance. We also conducted comprehensive ablation studies to demonstrate the efficacy of each component in our model design.
\end{itemize}

\section{Related Works}
\label{sec:relatedwork}

\noindent\textbf{Query-based 3D Object Detection.} 3D object detection plays a critical role in autonomous driving perception systems. Early multi-view camera-based methods~\cite{LSS, BEVFormer, BEVFusion} relied on explicit view transformation or implicitly learned dense BEV features via Transformers. Recent research~\cite{PETR, Sparse4D, Sparse4dV2, Sparse4dV3, CMT, StreamPETR} has explored sparse query techniques that offer significant advantages by focusing on foreground information. This approach improves computational efficiency and detection accuracy by allocating resources to regions likely containing objects of interest rather than processing the entire scene uniformly.
PETR~\cite{PETR} initializes object queries using 3D reference points that interact with 2D image features, directly learning spatial mappings without explicit perspective transformation. Sparse4D~\cite{Sparse4D} leverages 4D key points for sparse feature sampling, efficiently capturing object characteristics while filtering background noise.
CMT~\cite{CMT} introduces a multi-modal framework applying coordinate encoding across image and point cloud features, enabling focused attention on foreground objects while maintaining computational efficiency. StreamPETR~\cite{StreamPETR} implements object query propagation for temporal fusion, modeling object motion via motion-aware layer normalization and maintaining focus on moving objects while avoiding redundant processing of static backgrounds.

\noindent\textbf{Communication-efficient Cooperative Perception.} Cooperative perception\cite{liu2023towards, luo2025v2x, wang2025generative, li2024comamba, gao2025langcoop} enables connected and automated vehicles (CAVs) to exchange sensor data, significantly enhancing perception capabilities through mitigating occlusions, extending detection range, and improving autonomous driving safety and efficiency. This collaborative approach can be categorized into three primary fusion paradigms. Early fusion\cite{chen2019cooper, arnold2020cooperative} transmit raw sensor data among vehicles. While theoretically preserving maximum information, this paradigm imposes high bandwidth requirements. Late fusion \cite{melotti2020multimodal, fu2020depth, zeng2020dsdnet, shi2022vips, su2023uncertainty, su2024collaborative} share only high-level object detections or predictions, substantially reducing communication overhead but sacrificing cross-vehicle feature interaction capabilities, which leads to compromised performance in complex environmental conditions. Intermediate fusion\cite{V2VNet,CoBEVT,V2XViT, li2021learning, hu2022where2comm, qiao2023adaptive, wang2023core, gao2025stamp, huang2024actformer} has consequently emerged as the predominant research direction by effectively balancing accuracy and efficiency. This approach exchanges intermediate neural features (typically BEV feature maps) extracted from raw sensor inputs before executing final perception tasks. However, complete BEV feature maps contain substantial background information that contributes minimally to downstream perception tasks while consuming valuable bandwidth resources.

Several innovative approaches have been developed to further optimize communication efficiency within the intermediate fusion framework. Where2comm\cite{Where2comm} implements a spatial confidence-aware methodology that selectively transmits only the most critical feature information, significantly reducing bandwidth requirements. Similarly, CodeFilling\cite{CodeFilling} approximates feature maps using codebook-based representations and employs information filling techniques to select key information, achieving an optimal balance between communication efficiency and perceptual performance. Despite their effectiveness in bandwidth reduction, both methods requires completion of single-agent perception tasks prior to information transmission, resulting in increased computational latency. QUEST\cite{QUEST} represents an alternative approach that explores using object queries as information carriers in V2X scenarios, further reducing communication bandwidth requirements. However, QUEST’s restriction to two-agent collaboration severely limits the scalability of cooperative perception systems, making them impractical for real-world multi-agent traffic environments. In comparison, our proposed CoCMT framework enables efficient multi-agent collaboration while maintaining minimal communication bandwidth.

\label{sec:method}
\begin{figure*}[!t]
    \centering
    \includegraphics[width=0.9\textwidth]{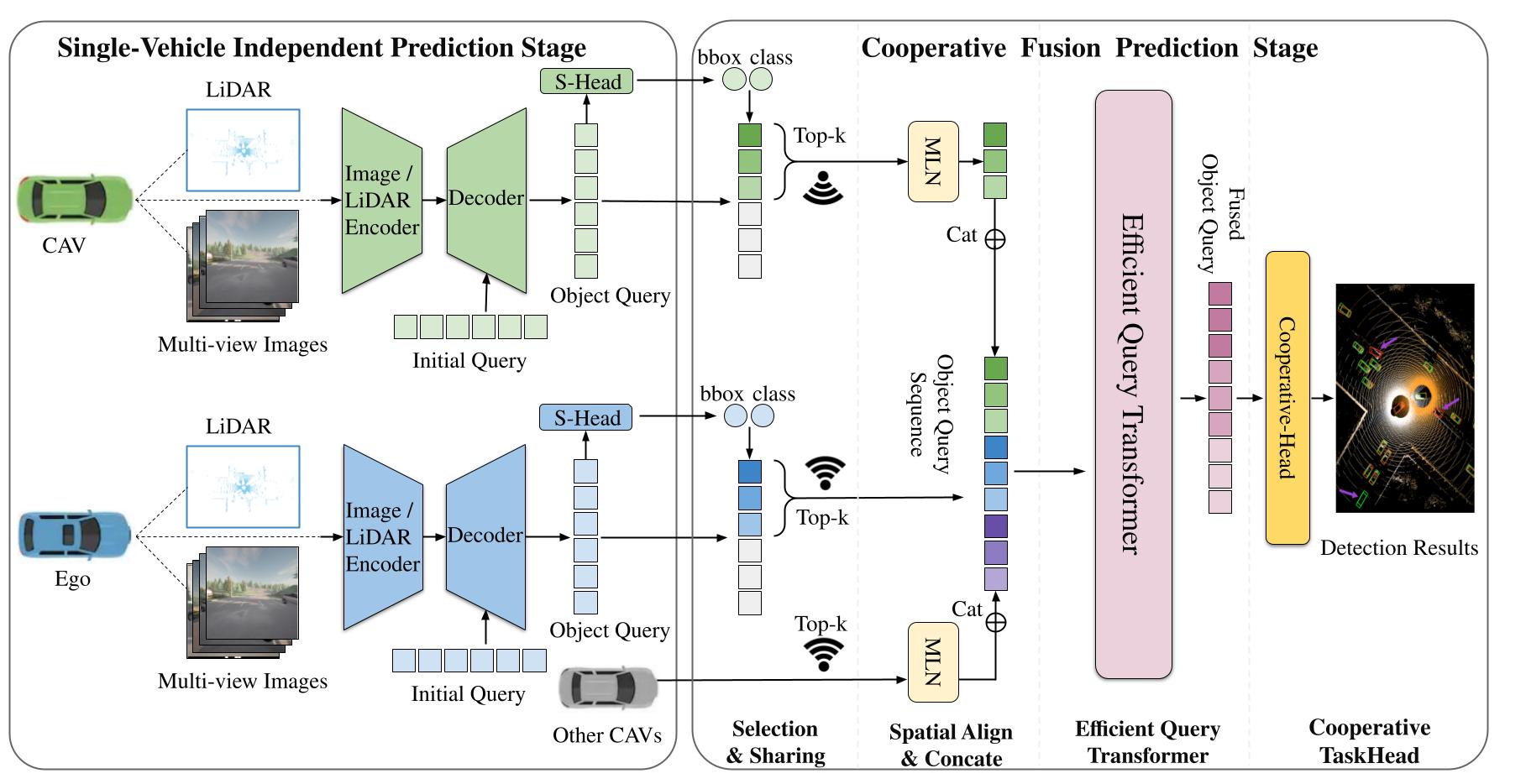}
    \vspace{-2mm}
    \caption{Overview of the CoCMT framework. The system operates in two stages: (1) single-agent independent prediction, which supports any query-based 3D object detection model with an S-head (single-agent task head), and (2) cooperative fusion prediction, comprising four components: Information Selection and Sharing, Spatial Alignment and Concatenation, the Efficient Query Transformer, and Cooperative Taskheads. MLN (Motion-aware Layer Normalization) handles spatial alignment for object queries.}
    \label{fig: Architecture Overview}
    \vspace{-3mm}
\end{figure*}

\section{Method}
We present CoCMT, illustrated in Figure ~\ref{fig: Architecture Overview}, divided into two stages: 1) the single-agent prediction stage, which we adopt the standard query-based learning objective to train the single-agent perception model, and 2) the cooperative fusion prediction stage. In the cooperative fusion prediction stage, we propose the \textbf{Efficient Query Transformer (EQFormer)} to restrict the interaction between object queries, achieved by applying several layers of attention masks.
To enhance the positive reinforcement between the two stages and aid the convergence of the framework, we propose a Synergistic \textbf{Deep Supervision mechanism (DSM)} that provides deep supervision for both stages simultaneously.
%

\subsection{Single-Agent Independent Prediction Stage}
In the first stage, we employ a query-based 3D object detection model to extract object queries, denoted as $ Q_{i} \in \mathbb{R}^{N \times D} $, where $ Q_{i} $ represents the set of object queries extracted from agent $i$. Each agent generates $ N $ queries with $ D $-dim features. We select $ Q_{i} $ as the core intermediate features in the cooperative fusion stage.
Notably, unlike most cooperative perception models that rely solely on the backbone for feature extraction, our approach retains the task heads of the model at this stage.
This retention allows us to incorporate additional predictive information—specifically, the 3D object centers  $C_i \in \mathbb{R}^{N \times 3}$  and object class scores  $S_i \in \mathbb{R}^{N \times C}$ —into the subsequent cooperative fusion prediction stage. By leveraging  $C_i$  and  $S_i$  alongside  $Q_i$, we enhance the effectiveness of the cooperative fusion by utilizing richer single-agent predictive outputs.
%
\subsection{Cooperative Fusion Prediction Stage}

\noindent \textbf{Information Selection and Sharing.}
Most query-based 3D object detection models initialize a large set of object queries to improve query coverage and accelerate model training~\cite{PETR,DN-DETR,CMT}.
However, during training, only a small portion of the object queries maintain strong associations with actual target objects, while the majority are predicted as background.
These background object queries do not contribute significantly to detection performance yet consume substantial transmission bandwidth when shared among agents.
To address this issue, we apply a Top-$k$  strategy to the object queries $Q_i$ based on the object classification scores $S_i$ output from the previous stage.
To balance effective fusion with reduced communication costs, we set $k$ to the maximum number of detectable objects by the connected and automated vehicles (CAVs).
After filtering, each CAV shares object queries $Q_i$, object centers $C_i$, and object class scores $S_i$. Additionally, the LiDAR poses of the CAVs are shared for subsequent spatial alignment.

\noindent \textbf{Spatial Alignment and Fusion.}
Due to the spatial differences between the CAVs and ego, their object queries exhibit significant spatial discrepancies.
To this end, we apply the Motion-aware Layer Normalization (MLN)~\cite{StreamPETR} to spatially align object queries.
Specifically, in our method, we first encode the transformation matrix $E_{cav}^{ego}$ from the CAV to the ego vehicle and then apply an affine transformation to $Q_{cav}$.
The object centers $C_{cav}$ of the CAVs are transformed into the ego vehicle's coordinate using $E_{cav}^{ego}$. After spatial alignment, we concatenate $Q_{ego}$ and $Q_{cav}$ for further fusion operations: $Q_{all}=Q_{ego}+\sum_iQ_{cav_i}$
To handle the dynamic number of connected vehicles in different V2V scenarios, we set the maximum number of connected vehicles in the system to $L$ and zero-padding the final query to maintain a fixed dimension of $L \times N$.

\noindent  \textbf{Efficient Query Transformer.}
After obtaining the object query sequences $Q_{all}$,  we input them into our Efficient Query Transformer (EQFormer).
EQFormer consists of three query-based self-attention blocks and utilizes the $M_{\text{all}}$ attention mask to enable focused and efficient interactions for object queries.
$M_{\text{all}}$ is a combination of three masking mechanisms specifically designed to address the challenges of object query fusion. Further details of the EQFormer are discussed in Section~\ref{ssec:eqformer}.

\noindent \textbf{Cooperative Task Head.} 
The fused object query sequence $Q_{fused}$, processed by the EQFormer, is fed into the task head for the 3D bounding box and object class prediction.
We normalize the object center sequences $C$ as reference points. Then, a bipartite matching algorithm~\cite{carion2020end} is applied to assign the predicted results to ground truth.
The details of the loss function are further explained in Section~\ref{ssec:sds}.

\subsection{Efficient Query Transformer}
\label{ssec:eqformer}

\begin{figure*}[t]
    \centering
    \includegraphics[width=0.9\textwidth]{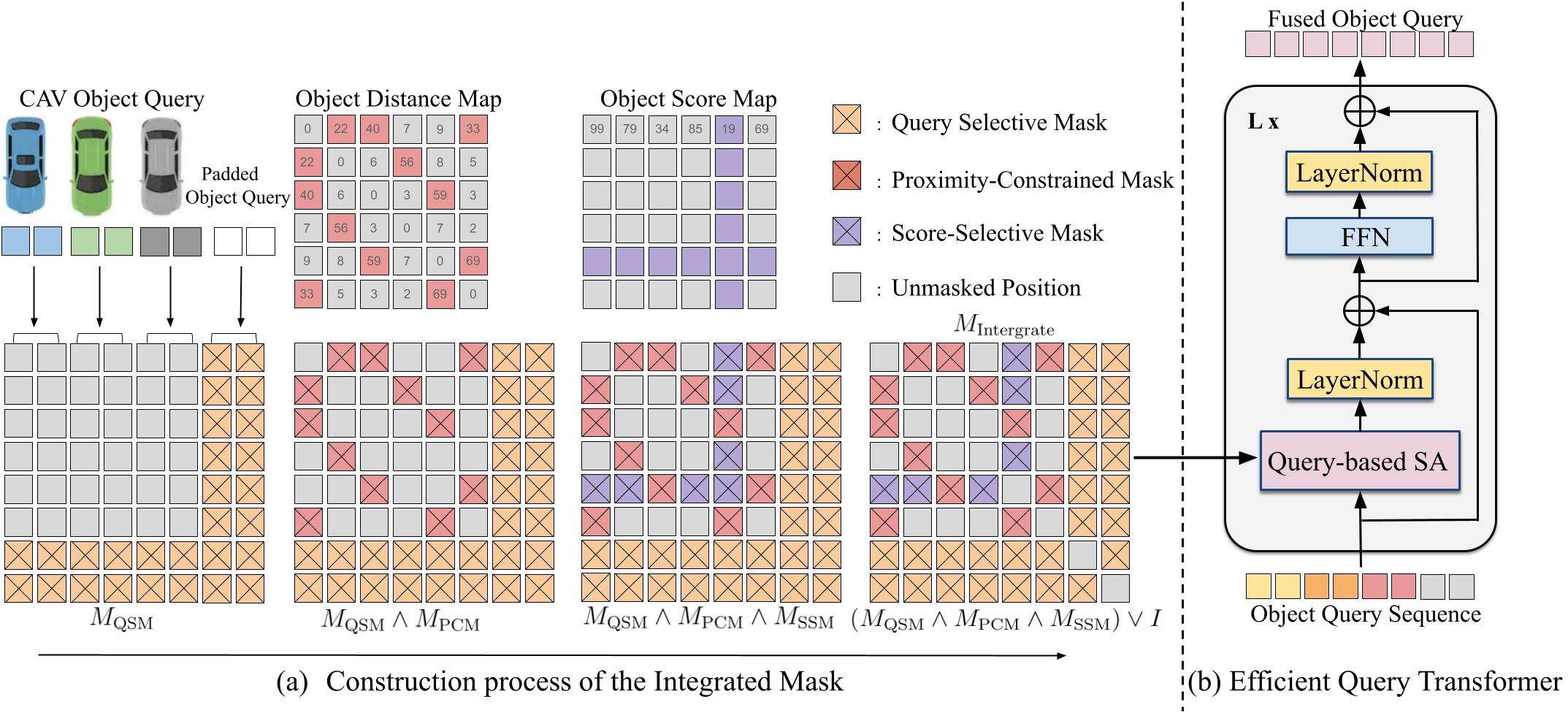}
    \vspace{-2mm}
    \caption{EQFormer architecture. Figure (a) illustrates the construction process of the integrated mask $M_{\text{all}}$. It consists of three mask mechanisms specifically designed to address the challenges of object query fusion: Query Selective Mask, Proximity-Constrained Mask, and Score-Selective Mask. Figure (b) shows the composition of the query-based self-attention block in EQFormer, which contains query-based self-attention equipped with $M_{\text{all}}$ and a feed-forward network (FFN).}
    \label{fig:efficient query transformer}
    \vspace{-3mm}
\end{figure*}

To address challenges in object query fusion, we propose the \textbf{Efficient Query Transformer (EQFormer)}, as shown in Fig.~\ref{fig:efficient query transformer}, introducing the integrated mask \( M_{\text{all}} \), which combines three masking mechanisms:
\begin{itemize}[leftmargin=10pt]
    \item Query Selective Mask (QSM): Prevents padded, invalid queries from interfering with interactions.
    \item Proximity-Constrained Mask (PCM): Limits interactions to spatially close object queries, mitigating failures due to contextual differences.
    \item Score-Selective Mask (SSM): Excludes background queries by leveraging object class scores.
\end{itemize}
We integrate \( M_{\text{all}} \) into a query-based self-attention block within the Multi-Head Self-Attention (MHSA) mechanism, combined with a Feed-Forward Network. EQFormer stacks three such blocks to efficiently fuse object query sequences.

\noindent \textbf{Query Selective Mask.}  
To ensure only valid queries participate in interactions, we introduce QSM \( M_{\text{QSM}} \in \mathbb{R}^{(L \times N)^2} \), which masks out zero-padded object queries. The matrix is defined as follows:
\begin{equation}
\resizebox{0.9\linewidth}{!}{$
M_{\text{QSM}}[i, j] = 
\begin{cases} 
0 & \text{if } 0 \leq i < AN \text{ and } 0 \leq j < AN \\
1 & \text{otherwise}
\end{cases}$}
\end{equation}
where $A$ represents the total number of CAVs in the current scene, $N$ denotes the number of queries per vehicle, thus $AN$ represents the total number of valid queries. Positions beyond $AN$ are assigned a value of 1, indicating masked object queries that are excluded from interactions, ensuring only valid queries are involved.

\noindent \textbf{Proximity-Constrained  Mask.}
To ensure that only spatially relevant object queries engage in the fusion stage, we introduce the Proximity-Constrained Mask (PCM).
This mechanism limits interactions based on the spatial proximity of the object centers corresponding to each object query.
This can potentially cause confusion during feature fusion when object centers are too far apart, and the contextual features between the corresponding queries may vary significantly.
To address this, PCM applies a distance threshold $\tau$ to restrict interactions. Specifically, let $C_{all} = \{c_1, c_2, \dots, c_{L \times N}\}$ represent object centers sequence, where $c_i$ denotes the object center corresponding to the $i$-th object query.
We define the spatial distance matrix $D$, with the element $D_{ij}$ representing the Euclidean distance between the $i$-th and $j$-th object centers, formulated as: $D_{ij} = \|c_i - c_j\|$.
Based on the matrix $D$ and the distance threshold $\tau$, we introduce the matrix of Proximity-Constrained Mask, expressed as follows:
\begin{equation}
\resizebox{0.9\linewidth}{!}{$
M_{\text{PCM}}[i, j] = 
\begin{cases} 
0, & \text{if } D_{ij} \leq \tau \\ 
1, & \text{if } D_{ij} > \tau 
\end{cases}
, \quad M_{\text{PCM}} \in \mathbb{R}^{(L \times N)^2}$
}
\end{equation}
Here, the values in the spatial distance matrix exceed the threshold $\tau$, $M_{\text{PCM}}$, which are set to 1, indicating that the corresponding object queries are masked. Conversely, $M_{\text{PCM}}$ are set to 0, allowing participation in interaction.

\noindent \textbf{Score-Selective Mask.}
In the Information Selection and Sharing module, we employed a Top-k filtering strategy to eliminate a significant number of object queries predicted as background.
To further exclude queries predicted as background during the fusion process and restrict interactions to queries associated with objects, thereby enhancing fusion efficiency, we introduced the Score-Selective Mask, an object-class score-based masking mechanism.
Specifically, let $S_{all} = \{s_1, s_2, \dots, s_{L \times N}\}$ represent the object class score sequence, where $s_i$ denotes the object class score of the $i$-th object query. Using the confidence threshold $\theta$, the matrix of the Score-Selective Mask is expressed as follows:
\begin{equation}
\resizebox{0.9\linewidth}{!}{$
M_{\text{SSM}, i} = 
\begin{cases} 
0, & \text{if } s_i > \theta \\
1, & \text{if } s_i \leq \theta
\end{cases}
, \quad M_{\text{SSM}} \in \mathbb{R}^{(L \times N)^2}$}
\end{equation}
Here, the confidence threshold $\theta$ is set to 0.20, consistent with the threshold used in post-processing. This choice effectively filters out most queries predicted as background while retaining a larger number of object-related queries for integration during the fusion process.
If the object score $s_i$ is less than or equal to $\theta$,  $M_{\text{SSM}}$ are set to 1, indicating that the corresponding object query is masked. Conversely, $M_{\text{SSM}}$ are set to 0, allowing the corresponding object query to participate in the fusion stage.

\noindent \textbf{Query-based Self-Attention Block.}
We integrate the above three object query interaction mechanisms into a unified mask,  $M_{\text{all}}$, which serves as the Attention Mask input for the self-attention block.
This self-attention block and feed-forward network (FFN), form our query-based self-attention block. 
These operations are formulated as follows:
\begin{equation}
M_{\text{all}} = (M_{\text{QSM}} \land M_{\text{PCM}} \land M_{\text{SSM}}) \lor I
\end{equation}
\begin{equation}
\resizebox{0.9\linewidth}{!}{$
\text{Attention}(Q, K, V, M_{\text{all}}) = \text{softmax}\left( \frac{Q K^T}{\sqrt{d_k}} + M_{\text{all}} \right) V
$}
\end{equation}
\begin{equation}
Q_{\text{fused}} = \text{EQFormer}(Q_{\text{all}}, M_{\text{all}}).
\end{equation}
The object query sequences $Q_{\text{all}}$ are fed into the EQFormer, achieving efficient fusion of object queries from different CAVs, and output the fused object queries $Q_{\text{fused}}$.

\subsection{Synergistic Deep Supervision}\label{Supervision}
\label{ssec:sds}
In current cooperative perception systems, improving the accuracy of a single agent's perception enhances the overall performance of the cooperative perception.
This implies a positive reinforcement between the Single-Agent Independent Prediction and the Cooperative Fusion Prediction Stages. To capitalize on this synergy, we introduce a Synergistic Deep Supervision approach, applying deep supervision to both stages simultaneously.
During the Single-Agent independent prediction stage, $Q_{\text{ego}}(i)$ from each layer of the ego vehicle's decoder is fed into the Single-TaskHeads.
In the collaborative fusion prediction stage, $Q_{\text{fused}}(j)$ from each layer of the EQFormer is fed into the Co-TaskHeads for regression and classification prediction. These operations are formulated as follows:
\vspace{-2mm}
\begin{equation}
    \hat{r}_{\text{sin}}(i), \hat{c}_{\text{sin}}(i) = \text{Sin-TaskHeads}(Q_{\text{ego}}(i))
\end{equation}
\begin{equation}
    \hat{r}_{\text{co}}(j), \hat{c}_{\text{co}}(j) = \text{Co-TaskHeads}(Q_{\text{fused}}(j)),
\end{equation}
where $\hat{r}_{\text{sin}}(i)$ and $\hat{r}_{\text{co}}(j)$ represent the regression predictions at each stage, while $\hat{c}_{\text{sin}}(i)$ and $\hat{c}_{\text{co}}(j)$ denote the classification predictions. 
We use Cross-Entropy Loss for object classification$\mathcal{L}_{\text{sin}}$ and $L_1$ Loss for bounding box regression$\mathcal{L}_{\text{co}}$, resulting final loss function:
\begin{equation}
\mathcal{L} = w_{\text{sin}} \mathcal{L}_{\text{sin}} + w_{\text{co}} \mathcal{L}_{\text{co}},
\end{equation}
where $w_{\text{sin}}$ and $w_{\text{co}}$ are weighting factors that balance the contributions of two losses.

\section{Experiments}

\begin{table*}[t]
\centering
\footnotesize
\setlength{\tabcolsep}{8.2pt}
\caption{Main performance and bandwidth comparison on OPV2V and V2V4Real Dataset. To further enhance model performance, we expanded the detection range of HMViT, PyramidFusion, and CoCMT to $[-102.4m, +102.4m]$ in the V2V-C setting of the OPV2V dataset. For CoCMT, we transmits the Topk(k=50) object queries during inference.}
\label{tab:main_performance_with_bandwidth}
\vspace{-2mm}
\begin{tabular}{l|cc|cc|cc|cc|c}
\toprule
Dataset & \multicolumn{6}{c|}{OPV2V} & \multicolumn{2}{c}{V2V4Real} \\ \midrule
Setting & \multicolumn{2}{c|}{V2V-C} & \multicolumn{2}{c|}{V2V-L} & \multicolumn{2}{c|}{V2V-H} & \multicolumn{2}{c|}{V2V-L} & \multirow{2}{*}{\makecell{Bandwidth \\ (Mb)}} \\ \cmidrule(lr){1-9}
Metric & AP50 $\uparrow$ & AP70 $\uparrow$ & AP50 $\uparrow$ & AP70 $\uparrow$ & AP50 $\uparrow$ & AP70 $\uparrow$ & AP50 $\uparrow$ & AP70 $\uparrow$ \\ \midrule
AttFuse\cite{OPV2VATTFuse} [ICRA 2022] & 0.447 & 0.184 & 0.895 & 0.779 & 0.624 & 0.411 & 0.701 & 0.454 & 536.8 \\
CoBEVT\cite{CoBEVT} [CoRL 2022] & 0.466 & 0.168 & 0.933 & 0.823 & 0.811 & 0.504 & 0.684 & 0.404 & 134.2 \\
V2X-ViT~\cite{V2XViT} [ECCV 2022] & 0.518 & 0.259 & 0.940 & 0.830 & 0.858 & 0.667 & 0.659 & 0.426 & 134.2 \\
HM-ViT~\cite{HMViT} [ICCV 2023] & 0.523 & 0.278 & 0.947 & 0.861 & 0.861 & 0.699 & 0.672 & 0.419 & 134.2 \\
HEAL~\cite{HEAL} [ICLR 2024] & \textbf{0.634} & 0.412 & 0.957 & \textbf{0.921} & 0.842 & 0.765 & \textbf{0.712} & 0.460 & 134.2 \\
\midrule
Late Fusion & 0.611 & 0.385 & 0.969 & 0.894 & 0.817 & 0.621 & 0.693 & 0.418 & 0.024 \\
CoCMT (ours) & \textbf{0.634} & \textbf{0.445} & \textbf{0.971} & 0.911 & \textbf{0.879} &\textbf{ 0.771} & 0.710 & \textbf{0.471} & 0.416 \\
\bottomrule
\end{tabular}%
\vspace{-3mm}
\end{table*}

\subsection{Datasets and Evaluation}

\noindent \textbf{Datasets.} 
We conducted extensive experiments on two multi-agent datasets: OPV2V~\cite{OPV2VATTFuse} and V2V4Real~\cite{V2V4Real}.
OPV2V~\cite{OPV2VATTFuse} is a large-scale, multi-modal simulated V2V perception dataset. The train/validation/test splits are 6,694/1,920/2,833, respectively.
V2V4Real~\cite{V2V4Real} is an extensive real-world cooperative V2V perception dataset,
which is split into 14,210/2,000/3,986 frames for training, validation, and testing, respectively.

\noindent \textbf{Evaluation.}
Following Xiang \emph{et al.} ~\cite{HMViT}, we evaluate three primary settings on this dataset: 1) Homogeneous camera-based detection (V2V-C), 2) Homogeneous LiDAR-based detection (V2V-L), and 3) Heterogeneous camera-LiDAR detection (V2V-H). We adopt Average Precision (AP) at Intersection-over-Union (IoU) thresholds of 0.5 and 0.7 to evaluate the model performance. The communication range between agents is set to 70m.

\subsection{Experimental Setups}

\noindent \textbf{Implementation Details.}
we employ the query-based 3D detection model, CMT~\cite{CMT}, as the primary model in the single-agent stage.
For the camera agent, we employ the CMT-C variant, which utilizes ResNet-50 as the camera encoder. For the LiDAR agent, we employ the CMT-L variant, which utilizes PointPillar as the LiDAR encoder. SPCONV2~\cite{spconv2} is applied for point cloud voxelization.
In both stages, all feature dimensions are set to 256, including point cloud tokens, image tokens, and object queries.

\noindent \textbf{Training strategy.} 
For V2V-L, we adopt the training strategy described in Section~\ref{Supervision},
We utilize a Top-$k$ selection strategy to transmit 120 object queries ($k=120$).
For V2V-H, we load the single-agent model (CMT-C) weights along with the multi-agent model weights trained in the V2V-L scenario.
The Top-$k$ selection strategy is applied to transmit $k=120$ object queries. 
For V2V-C, we train the model in an end-to-end manner, transmitting all 900 object queries.

\noindent \textbf{Compared Methods.}

We adopt \emph{late fusion} from the single-agent model of our framework as the baseline, which aggregates detection results from all CAVs and generates the final output.
For the intermediate fusion methods, we benchmark five SOTA methods: ATTFuse~\cite{OPV2VATTFuse}, CoBEVT~\cite{CoBEVT}, V2X-ViT~\cite{V2XViT}, HMViT~\cite{HMViT}, and HEAL (PyramidFusion)~\cite{HEAL}. These approaches all use feature maps as the medium for information exchange and employ LSS~\cite{LSS} to construct BEV features for camera branch.
In our experiments, ResNet50 and PointPillar served as the backbone networks for the camera and LiDAR branches, respectively.

\subsection{Quantitative evaluation}

\textbf{Perception performance and bandwidth.}
Figure~\ref{fig:opv2v_3branch_apvsbandwidth} demonstrates the trend of AP\@70 as a function of bandwidth on the OPV2V dataset. 
Under the V2V-L, V2V-C, and V2V-H settings, at the same bandwidth, our object-query-based model CoCMT significantly outperforms the feature-map-based intermediate fusion models. 
Additionally, as the bandwidth decreases, the performance degradation of the CoCMT is considerably smaller compared to the feature-map-based model, highlighting the transmission efficiency of object query and their adaptability to bandwidth limitations.
Table.~\ref{tab:main_performance_with_bandwidth} presents a performance comparison on the OPV2V and V2V4Real datasets.
Our proposed CoCMT model transmits only the Top-$k$ ($k=50$) object queries during inference, requiring just \textbf{0.416} Mb of bandwidth, which reduces bandwidth consumption by \textbf{83x} compared to the feature-map-based SOTA intermediate fusion model.
Despite the significant reduction in bandwidth, CoCMT still demonstrates excellent performance across multiple settings: on the OPV2V dataset, AP70 outperforms the SOTA intermediate fusion model by 2.7 and 0.6 points in the V2V-C and V2V-H settings, respectively; AP50 improves by 1.4 points in the V2V-L setting; and AP70 increases by 1.1 points in the V2V-L setting of the V2V4Real dataset.
This indicates that CoCMT not only offers significant transmission efficiency but also maintains superior performance in low-bandwidth environments.
Furthermore, CoCMT’s intermediate fusion method significantly outperforms the single-agent late fusion method, particularly on the V2V4Real dataset, where AP70 and AP50 are improved by 5.3 and 1.7 points, respectively.
This further highlights the performance advantages of our object-query-based intermediate fusion method.

\begin{figure}[ht]
    \centering
    \includegraphics[width=0.3\textwidth]{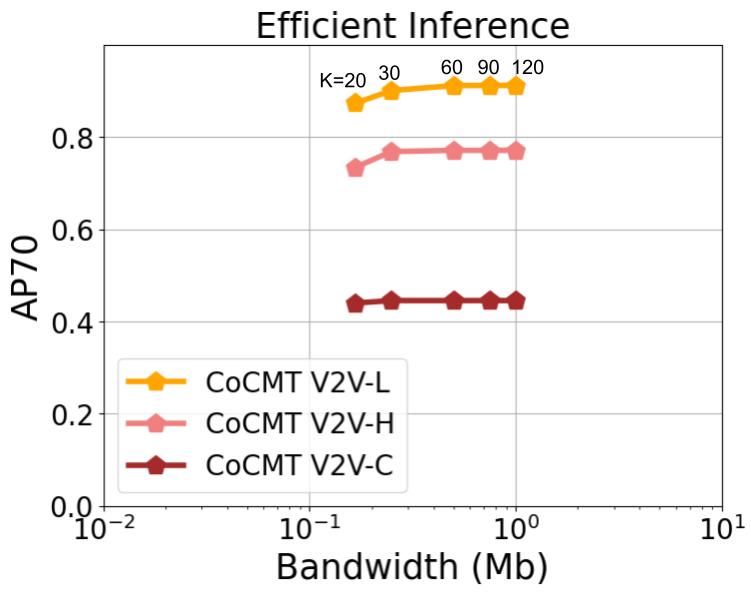}
    \caption{Top-$k$ selection strategy at inference with varying values of $k$ on the OPV2V-L dataset.}
    \label{fig:efficient_inference}
    \vspace{-3mm}
\end{figure}

\noindent\textbf{Efficient Inference Experiment.}
Figure.~\ref{fig:efficient_inference} demonstrates the performance variation of our model when reducing transmission bandwidth during inference.
Our model employs a class score-based Top-$k$ strategy during inference to reduce the number of transmitted object query, thereby lowering transmission bandwidth.
When the number of transmitted object query is reduced from 120 to 30, model performance remains nearly unaffected. Only when the transmission is reduced to 20, a slight performance drop is observed in the V2V-H and V2V-L settings.
This indicates that our object score mask effectively limits interactions to only strongly related object query.

\begin{figure*}[t]
    \vspace{-3mm}
    \centering
    \includegraphics[width=0.9\textwidth]{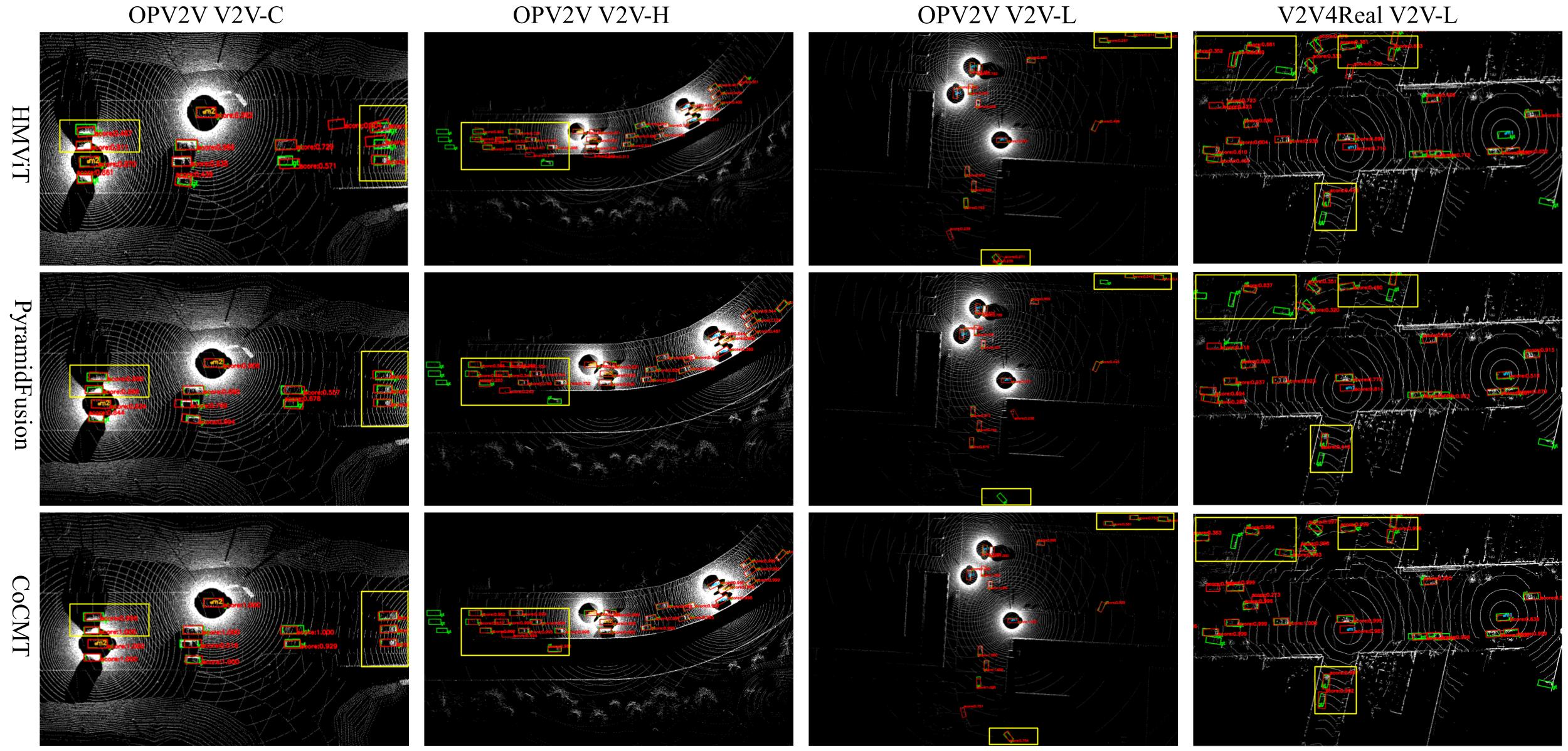}
    \caption{\textbf{Qualitative visualizations on the OPV2V and V2V4Real datasets.} Green and red 3D bounding boxes represent the ground truth and predictions, respectively. Key areas are highlighted with yellow boxes. Our method provides more accurate detection results and identifies more targets.}
    \label{fig:detection visualization}
    \vspace{-3mm}
\end{figure*}

\subsection{Alation Study}

\textbf{Component Ablation Study.}
We conducted ablation studies on the core designs of CoCMT, as shown in Table~\ref{tab:components_ablation}.
The results indicate that each design component significantly enhances model performance.
First, the Query Selective Mask $M_{\text{QSM}}$  filters out padded zero-value queries, preventing them from interfering with the fusion process and ensuring model stability.
Second, the Proximity-Constrained  Mask $M_{\text{PCM}}$ restricts object query interactions to spatially adjacent areas, enabling efficient and accurate fusion within a reasonable spatial range.
Lastly, the Score-Selective Mask $M_{\text{SSM}}$ further improves the focus of the fusion process by excluding queries predicted as background and restricting interactions to queries associated with objects. Combining these three masking mechanisms allows EQFormer to fuse object queries effectively for optimal performance.


\begin{table}[h]
\centering
\footnotesize
\setlength{\tabcolsep}{4pt}
\caption{Components ablation studies.}
\label{tab:components_ablation}
\begin{tabular}{cccc|cc}
\toprule
EQFormer & $M_{\textbf{QSM}}$ & $M_{\textbf{PCM}}$ & $M_{\textbf{SSM}}$ & AP50 $\uparrow$ & AP70 $\uparrow$ \\
\midrule
&          &              &          & 0.622 & 0.345 \\
\checkmark &               &               &               & 0.691 & 0.437 \\
\checkmark & \checkmark    &               &               & 0.691 & 0.440 \\
\checkmark & \checkmark    & \checkmark    &               & 0.721 & 0.465 \\
\checkmark & \checkmark    & \checkmark    & \checkmark    & 0.710 & 0.471 \\
\bottomrule
\end{tabular}%
\end{table}

\begin{wraptable}[9]{r}{0.22\textwidth}
\footnotesize
\setlength{\tabcolsep}{3pt}
\centering
\caption{$M_{\text{PCM}}$ distance ablation study results.}
\label{tab:distance_ablation}
\vspace{-2mm}
\begin{tabular}{c|cc}
\toprule

\multicolumn{1}{c|}{$M_{\text{PCM}}$} & \multicolumn{1}{c}{AP50 $\uparrow$} & \multicolumn{1}{c}{AP70 $\uparrow$} \\ 
\midrule

$+\infty$          & 0.690 & 0.419 \\
30m       & 0.696 & 0.440 \\
20m          & 0.700 & 0.452 \\
10m         & \textbf{0.710} & \textbf{0.471} \\
5m          & 0.683 & 0.430 \\

\bottomrule
\end{tabular}%
\vspace{-3mm}
\end{wraptable}

\noindent\textbf{Proximity-Constrained Mask Distance Ablation.}
The distance threshold in the Proximity-Constrained Mask directly influences the interaction range between object queries, which in turn has a significant impact on model performance.
In Table ~\ref{tab:distance_ablation}, we conducted an ablation study to evaluate the effects of different threshold values.
When the threshold is set to infinity, meaning no proximity-constrained restrictions are applied to interactions between object queries (i.e., the Proximity-Constrained  Mask is not used), the model's performance significantly declines.
We believe this is due to the large contextual differences between object queries, which lead to failed feature fusion. In contrast, when the distance threshold is set to 10 meters, the model achieves optimal performance.
Although increasing the threshold further expands the interaction range, it also introduces unreasonable interactions between object queries that are too far apart, ultimately resulting in reduced model performance.
This demonstrates that the Proximity-Constrained Mask plays a key role in improving model performance by effectively controlling the interaction range between object queries.

\begin{wraptable}[5]{r}{0.18\textwidth}
\vspace{-4mm}
\centering
\footnotesize
\setlength{\tabcolsep}{4pt}
\caption{SDS ablation.}
\label{tab:deepsupervision_ablation}
\vspace{-2mm}
\begin{tabular}{c|cc}
\toprule
SDS & AP50 $\uparrow$ & AP70 $\uparrow$ \\
\midrule
               & 0.698 & 0.422 \\
\checkmark    & 0.710 & 0.471 \\
\bottomrule
\end{tabular}%
\end{wraptable}

\textbf{Synergistic Deep Supervision (SDS) Ablation.}
Table~\ref{tab:deepsupervision_ablation} indicates that when SDS design is excluded, the model's performance significantly deteriorates. We attribute this to the vanishing gradient problem caused by the model's deep architecture, which hampers its convergence. However, when this design is applied, the model’s performance improves significantly. This highlights that Synergistic Deep Supervision aids model convergence and enhances the positive reinforcement between stages, ultimately leading to better overall performance.

\subsection{Qualitative evaluation}

\textbf{Detection visualization.}
Figure ~\ref{fig:detection visualization} presents the detection visualizations of CoCMT and PyramidFusion on the OPV2V and V2V4Real datasets.
As shown in the V2V-C setting of OPV2V, our CoCMT achieves higher detection accuracy, with predicted bounding boxes showing a greater overlap with ground truths.
In the V2V-L setting of both OPV2V and V2V4Real dataset, CoCMT detects more dynamic objects, showcasing the efficiency of using object query as a medium for information transmission.
In the V2V-H setting of OPV2V, CoCMT also achieves higher accuracy and broader detection coverage within the detection range of connected camera agents, demonstrating that our approach can effectively handle both homogeneous and heterogeneous multi-agent perception tasks through a unified and concise architecture.
\section{Conclusion}
We introduce CoCMT, a communication-efficient collaborative perception framework that transmits object queries to significantly reduce bandwidth consumption. Our proposed EQFormer features three masking mechanisms for precise query interactions and adaptive fusion, while synergistic deep supervision across both stages improves trainability and performance. Extensive experiments on simulated and real-world datasets demonstrate CoCMT's superior performance compared to prior work while achieving orders-of-magnitude bandwidth savings, advancing practical collaborative perception for resilient transportation systems.

\bibliographystyle{ieeetr} 
\bibliography{CoCMT}

\clearpage

\begin{figure*}[ht]
    \section{Appendix}
    \vspace{5mm}
    \centering
        \renewcommand{\thefigure}{A\arabic{figure}}

    \renewcommand{\thefigure}{A\arabic{figure}}
    \includegraphics[width=0.80\textwidth]{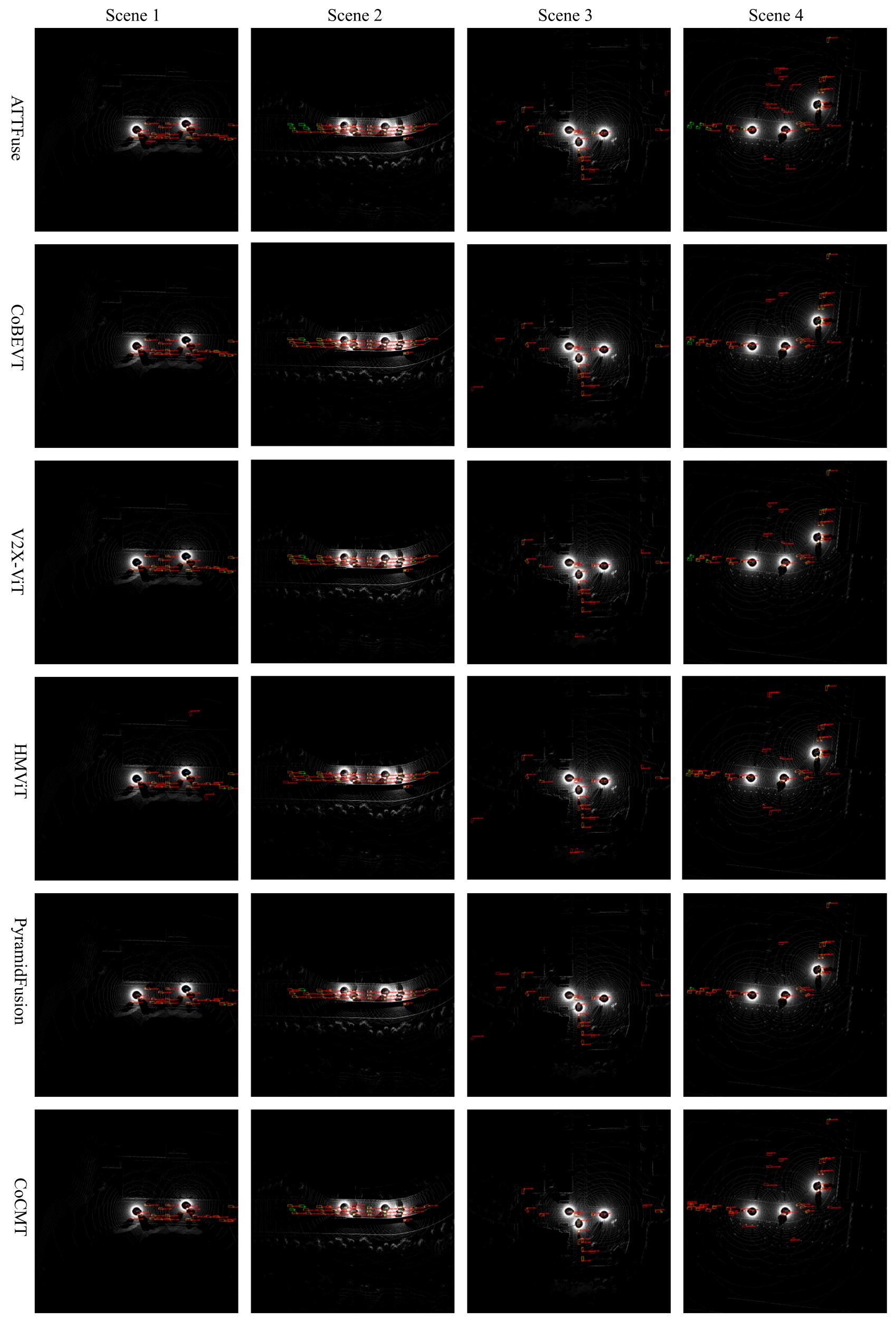}
    \caption{\textbf{Qualitative comparison on scenarios 1-4 under V2V-L setting in the OPV2V dataset.} The green and red bounding boxes represent the ground truth and prediction, respectively. Our method detected more dynamic objects.}
    \label{fig:OPV2V V2V-L Detection Visualization}
    \vspace{-1mm}
\end{figure*}

\begin{figure*}[ht]
    \vspace{-3mm}
    \centering
        \renewcommand{\thefigure}{A\arabic{figure}}

    \includegraphics[width=0.80\textwidth]{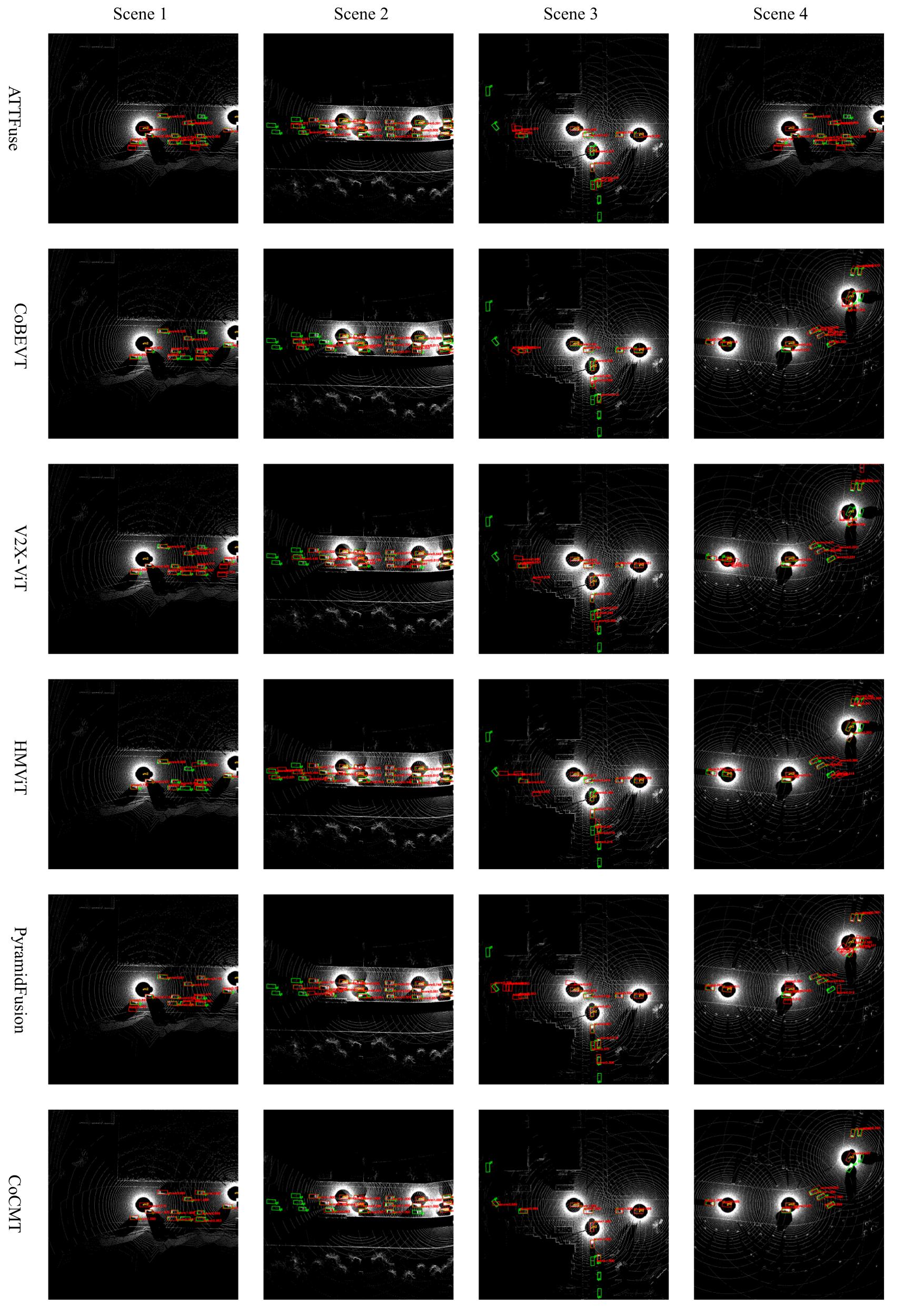}
    \caption{\textbf{Qualitative comparison on scenarios 1-4 under V2V-C setting in the OPV2V dataset.} The green and red bounding boxes represent the ground truth and prediction, respectively. Our method produced more accurate detection results.}
    \label{fig:OPV2V V2V-C Detection Visualization}
    \vspace{-1mm}
\end{figure*}

\begin{figure*}[ht]
    \vspace{-3mm}
    \centering
        \renewcommand{\thefigure}{A\arabic{figure}}
    \includegraphics[width=0.80\textwidth]{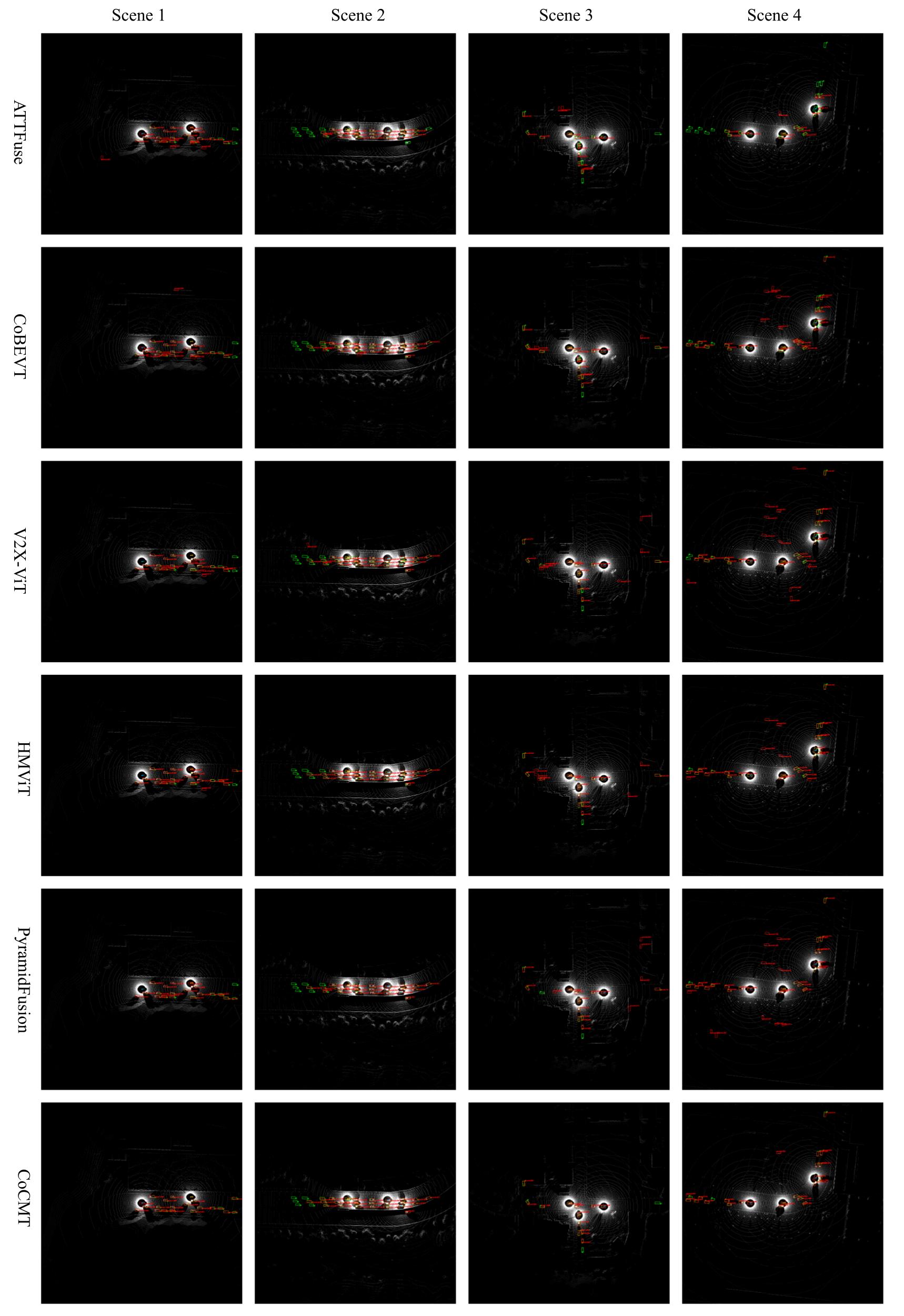}
    \caption{\textbf{Qualitative comparison on scenarios 1-4 under V2V-H setting in the OPV2V dataset.} The green and red bounding boxes represent the ground truth and predictions, respectively. Our method produced more accurate detection results and resulted in fewer false detection boxes.}
    \label{fig:OPV2V V2V-H Detection Visualization}
    \vspace{-1mm}
\end{figure*}

\begin{figure*}[t]
    \vspace{-3mm}
    \centering
        \renewcommand{\thefigure}{A\arabic{figure}}
    \includegraphics[width=0.90\textwidth]{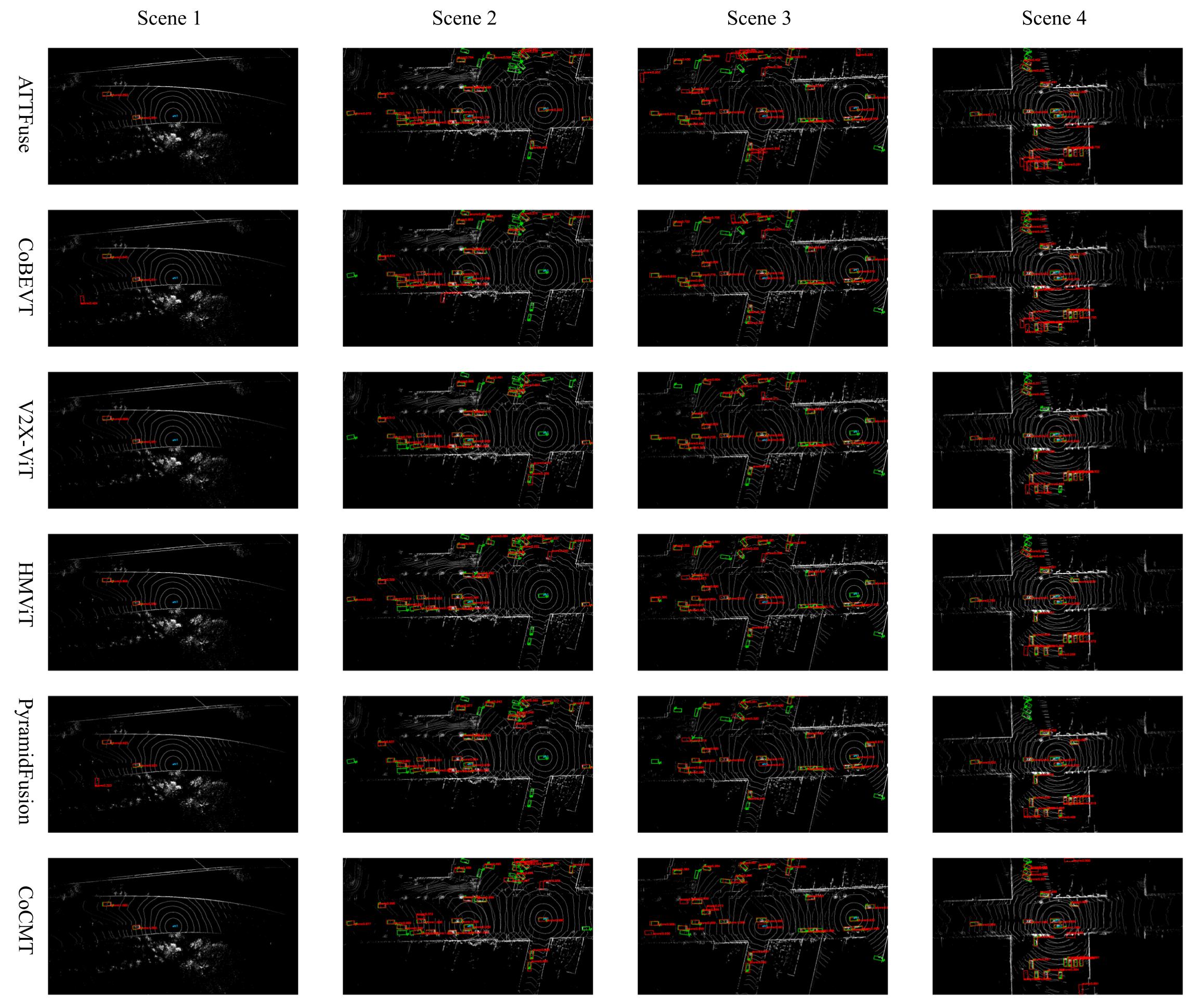}
    \caption{\textbf{Qualitative comparison on scenarios 1-4 in the V2V4Real dataset.} The green and red bounding boxes represent the ground truth and predictions, respectively. Our method produced more accurate detection results.}
    \label{fig:V2V4Real Detection Visualization}
    \vspace{-1mm}
\end{figure*}

\end{document}